\documentclass[sigconf, nonacm]{acmart}
\usepackage{xspace}
\usepackage{enumitem}
\usepackage{booktabs}
\usepackage[T1]{fontenc}
\usepackage[utf8]{inputenc}

\AtBeginDocument{%
  }
\begin{document}
\title{Efficient LLM-based Advertising via Model Compression and Parallel Verification}

\author{Wenxin Dong, Chang Gao, Guanghui Yu, Xuewu Jiao$^\dagger$, Mingqing Hu, Qiang Fu, Peng Xu, Penghui Wei, Hui Xu, Yue Xing, Shuanglong Li, Lin Liu}
\affiliation{%
  \institution{Baidu Inc.}
  \city{Beijing}
  \country{China}
}

\newcommand{\ProblemName}{{Generative targeting}\xspace}
\newcommand{\TechName}{{Efficient Generative Targeting}\xspace}
\newcommand{\scnONE}{{ad creative generation}\xspace}
\newcommand{\scnTWO}{{targeted advertising}\xspace}
\newcommand\secref[1]{Sec.~\ref{#1}}
\newcommand\equref[1]{Eq.~(\ref{#1})}
\newcommand\algref[1]{Alg.~\ref{#1}}
\newcommand\lemmaref[1]{Lemma~\ref{#1}}
\newcommand\theref[1]{Theorem~\ref{#1}}
\newcommand{\fakeparagraph}[1]{\vspace{1mm}\noindent\textbf{#1.}}
\newcommand\figref[1]{Fig.~\ref{#1}}
\newcommand\tabref[1]{Table~\ref{#1}}
\newcommand{\ie}{\emph{i.e.},\xspace}
\newcommand{\eg}{\emph{e.g.},\xspace}
\newcommand{\etc}{\emph{etc.}\xspace}

\renewcommand{\shortauthors}{Dong et al.}

\begin{abstract}
Recent advances in large language models (LLMs) have shown significant impact in various domains such as online advertising. However, traditional methods often struggle with inefficiencies in inference latency and high computational costs. To solve these problems, we propose an innovative \textbf{model compression} framework that integrates layer-adaptive group-wise quantization and layer-wise semi-structured sparsity optimized for GEMV workloads. We also introduce a novel \textbf{prefix tree-based parallel verification} strategy to further enhance efficiency. Extensive experiments in two scenarios---\scnONE and \scnTWO---validate the effectiveness of our approach.
The implementation of our framework in a leading advertising platform has been deployed in production, achieving a speedup of over 1.8$\times$ in real-world experiments while maintaining competitive precision.
\end{abstract}

\maketitle
\footnotetext{$^\dagger$Corresponding author.}
\footnotetext{This work was conducted in 2024 and the system has been deployed in Baidu's advertising platform since then.}

\section{Introduction} \label{sec:intro}
Recently, the advent of LLMs has introduced a paradigm shift in recommendation systems, enabling the transition from traditional discriminative-based architectures to generative-based approaches\cite{DBLP:journals/corr/2022NIR,DBLP:conf/icml/2024actionlouder}.
These models leverage  the world knowledge and reasoning abilities embedded in LLMs, offering superior generalization capabilities and robustness.
\ProblemName refers to the end-to-end, real-time process of seamlessly integrating generative LLMs with online advertising scenarios, directly generating relevant ads based on user queries. By embedding ad indices within the model, this new technological paradigm imposes higher demands on the regular updates of the model. 

However, as model parameters expand and real-time demands escalate, the complexity of inference in LLM-enhanced recommendation systems has risen, resulting in a scalability-efficiency dilemma.
This conflict positions latency-sensitive scenarios (\eg advertising) as the critical performance bottleneck, necessitating solutions that effectively optimize the inference without compromising recommendation quality.
 
\noindent\textbf{Motivation and Challenge}.
Existing research has primarily focused on simply \textit{applying} LLMs to recommendation systems (RS), with an emphasis on precision-side optimizations such as training strategy upgrade \cite{DBLP:conf/recsys/2023TallRec, DBLP:journals/corr/2023RecInterpret, DBLP:journals/corr/2024LLMEnhanced4Rec, DBLP:conf/wsdm/2024LLMRec} and prompting innovation \cite{DBLP:conf/sigir/2024LlaRa,DBLP:conf/sigir/LLMandRank}. These approaches have demonstrated exceptional accuracy, significantly improving the predictive performance of recommendation systems.
However, these studies often overlook efficiency-side optimizations, which is particularly detrimental in time-sensitive scenarios like online real-time inference for commercial ad delivery.

\noindent\textbf{Our Main Idea}.
To tackle these technical challenges, we are motivated to propose a pioneering and effective solution to the \ProblemName problem.
Specifically, we innovatively designed an index-compressed data structure to support semi-structured layer-wise FP16 mixed sparsification and used adaptive group-wise quantization techniques to achieve model compression from FP16 to INT4. Additionally, we developed a custom kernel that supports mixed-precision computation with INT4 weight-only GEMV and mixed sparsity to enhance matrix acceleration efficiency.
Moreover, in business, we employed a hierarchical clustering algorithm\cite{DBLP:conf/nips/2022Cluster} to construct plaintext data into a semantically structured prefix tree. 
By dynamically evaluating the time difference between generating and verifying remaining tokens, we identify the optimal point for launching prefix tree-based parallel verification. This approach enables decoding the entire remaining sequence length in a single step, which can reduce decoding step and accelerate model inference.

\noindent\textbf{Contribution}.
The main contributions of our paper are summarized as follows:
\vspace{-1ex}
\begin{itemize}[leftmargin=*]
\item
We optimized Compressed Sparse Row (CSR) by proposing an index-compressed 2bit-CSR which is used for mixed sparsity, reducing the size of the index and weights to 30\% of the original CSR size before optimization.
\item 
We independently developed a custom SparseGemv acceleration kernel supporting INT4 with sparse matrix multiplication, filling the gap left by NVIDIA's sparse acceleration libraries (cuSparse/cuSparseLT) for efficient GEMV operations.
\item 
We innovatively proposed adaptive grouping quantization, enhancing the flexibility and precision of the quantization process.
\item 
We are, to our knowledge, the first to propose a complete workflow combining prefix tree-constrained decoding with beam search\cite{DBLP:conf/emnlp/2016BeamSearch}, applying it to the generative task in advertising targeting scenarios.
\item 
Extensive experimental data demonstrates that our optimization achieved an improvement of over 78\% in inference speed. The proposed system has been deployed in Baidu's advertising platform, serving real-time traffic at scale.
\end{itemize}


\vspace{-1ex}
\section{Related Work}\label{sec:related}
\subsection{LLM in Recommendation System}

The integration of Large Language Models (LLMs) into recommendation systems has emerged as a promising research direction. 
Current research primarily encompasses several key areas: (1) feature engineering and representation learning, where LLMs are leveraged to generate rich semantic embeddings \cite{DBLP:conf/www/LLMEmbedNoteLLM,DBLP:conf/recsys/LLMEmbGengRLP}; (2) ranking optimization, where LLMs improve result ordering accuracy \cite{DBLP:conf/sigir/LLMandRank, DBLP:conf/sigir/2024LlaRa,DBLP:journals/corr/2023RecRanker,DBLP:conf/www/2024ReLa}; and (3) user interaction enhancement, focusing on natural language interfaces and dialogue-based recommendations \cite{DBLP:conf/recsys/2023TallRec, DBLP:conf/emnlp/2023UUniCRS}.
A parallel advancement involves the adoption of generative recommendation\cite{DBLP:conf/icdm/2018attention_2,DBLP:journals/corr/2023GPTRec}, particularly the self-attention mechanisms. Notable work by \cite{DBLP:conf/icml/2024actionlouder} demonstrates that unified generative recommendation models have achieved breakthrough performance, surpassing traditional hierarchical deep recommendation systems. 
Furthermore, recent research has proposed efficient scaling solutions \cite{DBLP:conf/icml/tradition_scaleup/wukong} for traditional search and recommendation models.
However, despite these advances, current approaches struggle with high computational costs and latency, making them impractical for online inference scenarios with strict response time requirements. This necessitates new methods that balance performance and efficiency.

\subsection{LLM Inference Acceleration}
Recent advances in LLM inference acceleration have primarily focused on two orthogonal complementary approaches: model compression techniques and decoding optimization strategies.  

\fakeparagraph{Model Compression}  
Quantization and sparsity have emerged as a fundamental methodology for reducing model size. Seminal works including \cite{DBLP:conf/ram/GPTQ,DBLP:conf/mlsys/AWQ}, and their derivatives \cite{DBLP:conf/aaai/2024OWQ,DBLP:conf/iclr/2024OmniQuant} have established robust 4-bit quantization baselines.
Nevertheless, these methods predominantly employ pre-determined grouping strategies, which easily overlook the intrinsic heterogeneity in weight sensitivity distributions.  
In parallel, sparsity research has evolved from unstructured pruning to hardware-aware semi-structured paradigms. SparseGPT\cite{DBLP:conf/icml/SparseGPT} and Wanda\cite{DBLP:conf/iclr/2024Wanda} achieve sparse pruning, also show promise in reducing parameters while maintaining model quality. 
However, SparseGPT\cite{DBLP:conf/icml/SparseGPT} is limited by its layer-wise loss optimization and reliance on second-order derivatives, which constrain parameter adjustments. Retraining-based methods offer a broader optimization space, enabling better global performance. 

\fakeparagraph{Speculative Decoding}  
Speculative decoding accelerates autoregressive decoding by enabling parallel verification of candidate tokens, as proposed by \cite{DBLP:conf/icml/LeviathanSpec, DBLP:journals/corr/ChenSpec}. Innovations like Medusa\cite{DBLP:conf/icml/2024Medusa} and SpecInfer\cite{DBLP:conf/asplos/2024SpecInfer} utilize tree-structured attention for parallel prediction. However, existing methods apply parallel decoding uniformly, which can be inefficient. We introduce dynamic parallelism initiation, transitioning strategically from sequential to parallel decoding based on confidence and computational cost, optimizing both response time and accuracy for industrial recommendation systems.

\section{Solution} \label{sec:solution}

\subsection{Overview}
The system consists of two primary components: \textit{Model Compression} and \textit{Prefix Tree Parallel Verification}, designed to optimize efficiency and accuracy in generative targeting tasks.

\fakeparagraph{Model Compression} It focuses on reducing model size and computational overhead without sacrificing precision. It includes Group-wise Adaptive Quantization, which, based on parameter sensitivity, groups the layers of the model into sensitive layers and non-sensitive layers, with fine-grained (more groups) quantization for sensitive groups and coarse-grained (fewer groups) quantization for non-sensitive groups. The Layer Adaptive Sparsification component prunes redundant parameters across different layers based on their relative importance, applying varying ratios to ensure that critical layers retain more information while less significant layers are pruned more aggressively. These compression techniques are seamlessly integrated with the kernel for optimized execution.

\fakeparagraph{Prefix Tree Parallel Verification} is responsible for efficient and accurate token prediction. The process begins with Prefix Tree Construction, where hierarchical clustering\cite{DBLP:conf/nips/2022Cluster} transforms plain text into a structured prefix tree (trie), the trie structure is wide at the top and narrow at the bottom, with fewer candidate results as one goes deeper. The Parallel Verification Trigger dynamically evaluates and identifies optimal trigger points for verification, ensuring a balance between processing speed and accuracy. Finally, Tree-Based Parallel Verification reconstructs the prefix tree for sequence generation, utilizing parallel decoding to accelerate results while preserving precision. 


\subsection{Model Compression}
\fakeparagraph{Adaptive Group-Wise Quantization}
To enhance the flexibility and precision of the quantization process, we propose an adaptive group-wise method that tailors the quantization strategy to the sensitivity of individual linear layers. 
Drawing inspiration from the weight sensitivity analytical solution of the near-lossless methods\cite{DBLP:conf/iclr/2024SqQR}, we further integrated its measurement strategy into our approach as analytical standard.
Therefore, channels in the layer which identified as sensitive are allocated more groups to achieve finer quantization granularity, minimizing quantization errors.
Conversely, non-sensitive channels, which exhibit robustness to quantization-induced changes, are assigned fewer groups, reducing computational overhead without sacrificing accuracy.

\begin{figure}[H]
\centering
\includegraphics[width=\linewidth]{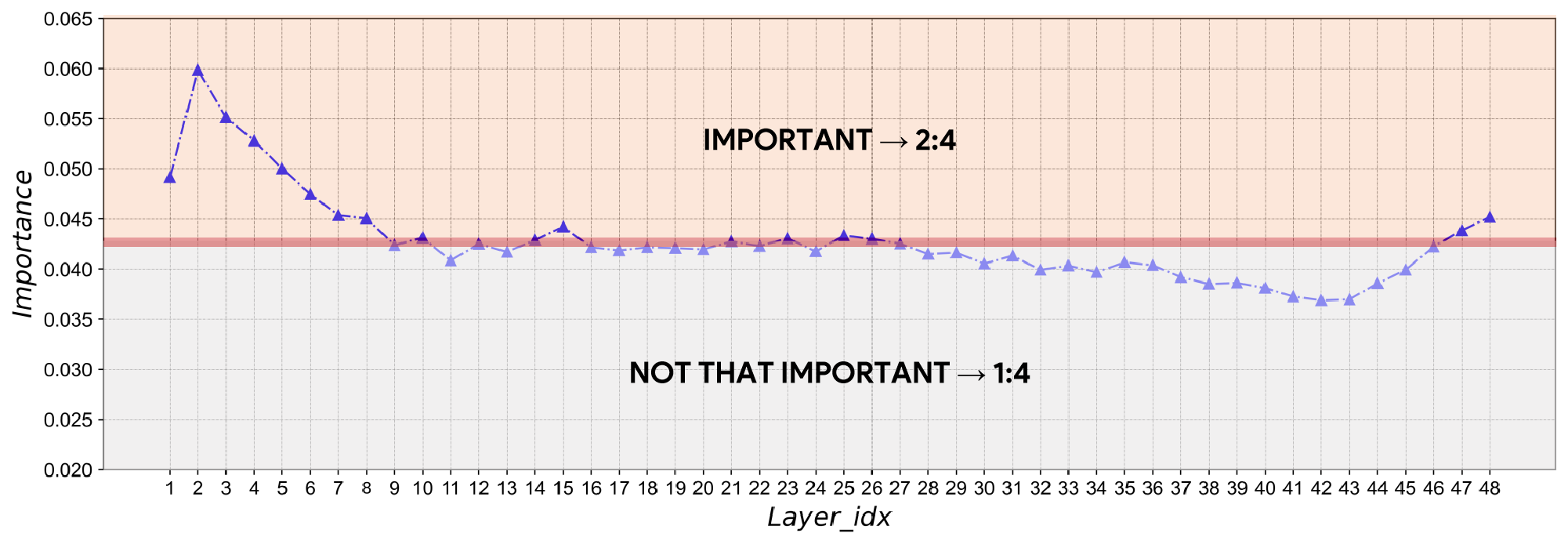}\vspace{-1.5ex}
\caption{Transformer Layer Importance}
\label{fig:layer_importance}
\vspace{-3ex}
\end{figure}

\fakeparagraph{Sparsification}
To further improve the inference efficiency while not compromising the precision, we propose a layer-wise adaptive $N:M$ pruning strategy. Based on semi-structured sparsity\cite{DBLP:conf/icml/SparseGPT}, we incorporate the importance-based weight selection. In essence, this approach enables layers to be pruned at varying degrees of sparsity, with more critical layers retaining a higher density (2:4) and less critical layers exhibiting greater sparsity (1:4), as shown in \figref{fig:layer_importance}.

\vspace{-1ex}
\vspace{-1ex}

Building upon the pruning criterion established in WandA\cite{DBLP:conf/iclr/2024Wanda}, we propose a refined methodology to quantify the importance of individual elements in the weight matrix. 
Let $\mathbf{w}_0 \in \mathbb{R}^d$ denote the dense weight vector prior to sparsification, and $\mathbf{w} = \mathbf{w}_0 + \delta \mathbf{w}$ represent its perturbed counterpart after pruning. The approximation error induced by the sparsification operation can be derived through a second-order Taylor expansion:
\begin{equation}
\delta E = E(\mathbf{w}) - E(\mathbf{w}_0) = \left( \frac{\partial E}{\partial \mathbf{w}} \right)^\top \delta \mathbf{w} + \frac{1}{2} \delta \mathbf{w}^\top \mathbf{H} \delta \mathbf{w} + O(\|\delta \mathbf{w}\|^3)
\end{equation}
where $\mathbf{H} = \frac{\partial^2 E}{\partial \mathbf{w}^2}$ is the Hessian matrix. The primal optimization objective is to find a set of weight elements to minimize the induced approximation error. Under the assumption of element-wise independence and ignoring higher-order terms, we formulate the Lagrangian constrained optimization problem: For an individual weight element $w_q$(satisfying $w_q + \delta w_q = 0$), the error term will be:
\vspace{-0.5ex}
\begin{equation}
\left( w_q \cdot \|\mathbf{x}_q\|_2 \right)^2 + w_q \cdot \left| \frac{\partial E}{\partial w_q} \right|
\end{equation}
where $\mathbf{x}_q$ represents the part of the input activation that will operate with $w_q$ and $E$ denotes the loss function.
Thus, the criterion for any element $\mathbf{W}_{ij}$ in the given weight matrix $\mathbf{W}$ becomes:
\begin{equation}
|\mathbf{W}_{ij}| \cdot \|\mathbf{X}_j\|_2 + |\mathbf{W}_{ij}| \cdot \left| \frac{\partial E}{\partial \mathbf{W}_{ij}} \right|
\end{equation}

\subsection{Revised Kernel}
Semi-structured sparse matrices necessitate supplementary index data, which inherently lead to extra bandwidth consumption and computational burden. 
Thus, we have mainly focused on tackling these pivotal challenges, accommodating the real-world scenarios. Based on our previous generation of Gemv (General Matrix-Vector Multiply) technology, we have systematically refined the technology, culminating in the development of a more efficient variant—SparseGemv. 

\fakeparagraph{Sparse Index Structure}
Drawing inspiration from the traditional matrix compression method, CSR, we devised an advanced data structure to further condense index information, minimizing the additional overhead of the memory bandwidth. 
Given a weight matrix, we extract its non-zero elements and record their column positions in a compact index format. Then we reordered the indices and converted them into binary representations sequentially. Finally, the binary indices are grouped and encoded into a compact format, such as hexadecimal, to significantly compress parameters and accelerate copying.

\fakeparagraph{Computational Acceleration}
Based on the aforementioned index storage structure, we implemented a sparse matrix multiplication kernel (SparseGemv) to accelerate computations involving sparse matrices. The objectives are two folds: 
\begin{itemize}[leftmargin=*]
    \item 
    \textit{Accelerate Memory Access}: First, Shared Memory ({SharedMem}) was employed to tackle performance degradation caused by dynamic access. Activations, originally stored in Local Memory due to runtime-determined indices, were moved to Shared Memory to improve performance. 
    However, direct loading of activations caused bank conflicts. To resolve this, a shuffle-based data layout was implemented, which reassigned threads to non-conflicting memory banks by staggering intervals, eliminating conflicts and enabling efficient memory operations.
    Additionally, block-level data sharing was introduced to enhance parallelism. Multiple warps collaboratively loaded activations from High-Bandwidth Memory (HBM) into Shared Memory, improving data transfer efficiency.
    \item 
    \textit{Reduce Computational Burden}: For computation optimization, we proposed two methodologies: assembly instruction selection and index utilization. By utilizing the $lop3.b32$ assembly instruction for INT4-to-FP16 weight dequantization, the required instructions were reduced significantly, improving efficiency. Meanwhile, the Compressed Sparse Row (CSR) indexing method was optimized to reduce storage overhead while maintaining retrieval efficiency. Sparse indices were packed into $uint16_t$ variables, and reverse-order storage with bit operation accelerates index calculation, ensuring compactness and fast access.
\end{itemize}

\subsection{Tree-based Parallel Verification}


In business scenarios, we explored unique data pattern in generative targeting, devising efficient organization strategies to accelerate inference for specific business needs. To achieve this, we have built three core components within our workflow: \textit{Prefix Tree Construction}, \textit{Parallel Verification Trigger}, \textit{Tree-Based Verification}.

\fakeparagraph{Hierarchical Clustering for Prefix Tree Construction} 
This module aims to transform plain text into structured IDs and subsequently used to build a prefix tree that is structurally designed to be broad at the top and narrow at the bottom, which can cater to facilitating efficient decoding and verification. 
To achieve this objective, we initially collect all the advertiser-specific plaintext identifiers (\eg commercial entities like "Taobao" or "Baidu") as transformation inputs. 
Subsequently, we adapt the Differentiable Search Index(DSI) algorithm\cite{DBLP:conf/nips/2022Cluster}. Originally, the semantically structured identifiers are resolved by making the target corpus into $k$ clusters recursively, aligning with the intention to capture more information and effectively reduce search space after each decoding step. Similarly, we implemented this hierarchical clustering process during the trie construction. We treat the plain-text as corpus in the DSI model, with hyper parameters denoting the clusters category number and recursive demarcation, we can then ensure the precondition of validation is met.


\fakeparagraph{Parallel Verification Trigger}
During the process, it is vital for us to decide when to trigger the parallel verification. Intuitively, the switch starts when the predicted benefit is maximized. Exhaustive experimental results reveal the overhead of prefix tree parallel verification scales linearly with the length of candidate results, while the time required for a single autoregressive step remains approximately constant on the same hardware and model. 
This leads to our proposed solution: the trigger moment is dynamically determined based on the time difference between generating the remaining tokens and verifying them. It demonstrates that the system calculates the trade-off between the overhead and benefit of parallel verification in real-time during the inference process. Parallel prefix tree verification is triggered when the predicted benefit is maximized, ensuring efficient utilization of computational resources while maintaining accuracy.

\fakeparagraph{Tree-Based Parallel Verification}  
The tree-based parallel verification process was inspired by the speculative decoding ideas\cite{DBLP:conf/asplos/2024SpecInfer}. 
This method is a three-stage workflow involving restructuring the tree to construct input sequences, decoding them with tree-based constraints, and performing beam search to select optimal sequences
The decoding process leverages a trie and begins with a depth-first search (DFS) to construct input sequences and a tree mask\cite{DBLP:conf/icml/2024Medusa}, ensuring only valid paths are considered. Padding handles varying sequence lengths due to beam search. These sequences are processed in parallel by the decoder, which outputs a probability distribution $T_{score}$ for each position. Using the trie, $T_{score}$ is filtered to retain valid token scores, which are combined with parent node scores via logarithmic summation to compute node scores $B_{score}$.  
Beam search\cite{DBLP:conf/emnlp/2016BeamSearch} selects the top sequences based on $B_{score}$. At the final position, the highest-scoring nodes (up to $beam_{size}$) are traced back to reconstruct valid sequences. This approach ensures efficient and accurate decoding by applying structural constraints from the prefix tree while accelerating the process.





\section{Experiments} \label{sec:exp}
In this section, we select ERNIE 1.5B as our backbone model to evaluate our framework on two real-world scenarios: \scnONE and \scnTWO. 
All experiments are implemented using PaddlePaddle.
For \scnONE, experiments are conducted on an NVIDIA A10 GPU with a beam size of 1, while \scnTWO utilizes an NVIDIA A30 GPU with a beam size of 20, because more advertising candidates need to be selected.

\subsection{Results of Targeted Advertising}
\begin{figure}[htbp]
\centering
\begin{minipage}{0.48\linewidth} 
    \centering
    \includegraphics[width=\linewidth]{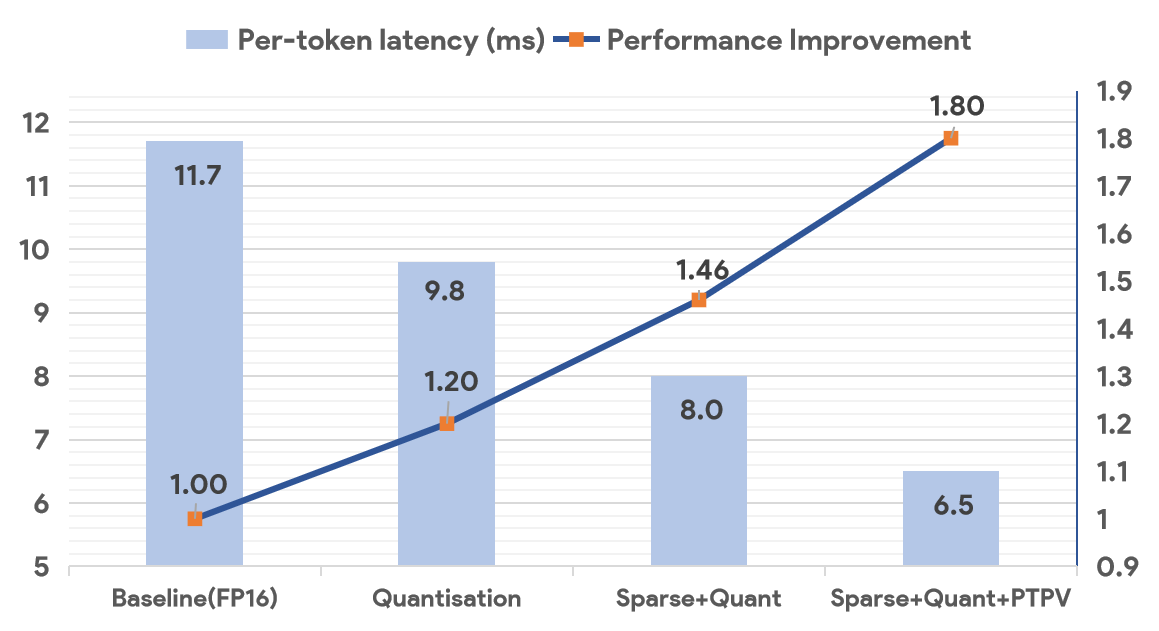}\vspace{-1.5ex}
    \label{fig:targeted_ads}
    \vspace{-3ex}
\end{minipage}
\hfill
\begin{minipage}{0.48\linewidth} 
    \centering
    \renewcommand{\arraystretch}{1.2} 
    \setlength{\tabcolsep}{3pt} 
    \scriptsize 
    \begin{tabular}{lcc} 
        \toprule
        Technique & AVG Len.  & RECALL \\
        \midrule
        Baseline (FP16) & 6  & 0.084 \\
        Quantization & 6  & 0.083 \\
        Sparse + Quant & 6  & 0.083 \\
        Sparse + Quant + PTPV & 6  & 0.081 \\
        \bottomrule
    \end{tabular}
    \label{tab:targeted ad}
    \vspace{-3ex}
\end{minipage}
\caption{Comparison on Targeted Advertising Scenario.}
\label{targeted_ads}
\end{figure}

To evaluate the efficiency of our optimization techniques, we conducted a series of experiments comparing different configurations of our framework. 
For datasets, we utilized a proprietary dataset consisting of internal commercial traffic data collected by our company. Due to its sensitive nature, the dataset is not publicly available.
For advertising evaluation metrics, we use latency and recall to assess the trade-off between inference acceleration and precision sacrifice. 
As shown in \figref{targeted_ads}, our individual techniques demonstrate notable improvements in performance, their combination ensures both significant speedup and acceptable recall levels, making the complete framework well-suited for practical online deployment. 

\vspace{-1ex}
\vspace{-3ex}
\subsection{Results of Ads Creative Generation}
\begin{figure}[htbp]
\centering
\begin{minipage}{0.48\linewidth} 
    \centering
    \includegraphics[width=\linewidth]{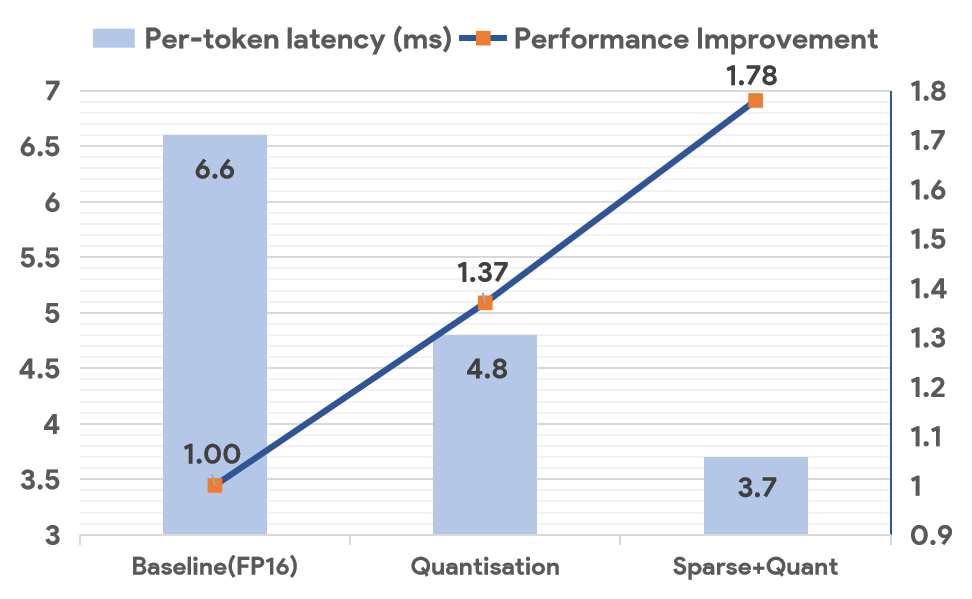}\vspace{-1.5ex}
    \label{fig:creative_generation_1}
    \vspace{-2.5ex}
\end{minipage}
\hfill
\begin{minipage}{0.48\linewidth} 
    \centering
    \renewcommand{\arraystretch}{1.2} 
    \setlength{\tabcolsep}{3pt} 
    \scriptsize 
    \begin{tabular}{lccc} 
        \toprule
        Technique & AVG Len. & Meteor & BLEU \\
        \midrule
        Baseline (FP16) & 17.5 & 0.6345 & 0.4247 \\
        Quantization & 17.6 & 0.6283 & 0.4178 \\
        Sparse + Quant & 17.5 & 0.6127 & 0.4038 \\
        \bottomrule
    \end{tabular}
    \label{table:creative_generation_2}
    \vspace{-2.5ex}
\end{minipage}
\vspace{-1.5ex}
\caption{Comparison on Ads Creative Generation Scenario}
\vspace{-1ex}
\label{creative_generation}
\end{figure}

The assessment of the \scnONE focus primarily
on performance with creative rewriting and keyword summarization.
We use \textit{CSL}\cite{DatasetCSL} as the dataset, which contains metadata from 396,209 core Chinese academic journal articles published between 2010 and 2020.
As demonstrated in \figref{creative_generation}, the baseline established the quality upper bound, while sparsification and quantization induce marginal decreases in accuracy yet yield significant efficiency gains. Notably, the hybrid approach achieves the best trade-off between efficiency and quality, making it operationally viable for practical deployments in real-world applications.

\subsection{Ablation Studies}
To evaluate the effectiveness of the optimization techniques in ernie 1.5B, we conduct an ablation study by progressively integrating different components. INT4 adaptive group quantization and layer-adaptive hierarchical sparsification are analyzed in terms of per-token latency with BLEU and Meteor scores. 
As shown in \tabref{ablation}, quantization significantly reduces latency while maintaining high BLEU and Meteor scores, demonstrating its effectiveness in preserving generation quality. In contrast, sparsification yielded only moderate efficiency gains. Conservative sparsification approaches($2:4$) show limited efficiency improvements, while aggressive sparsification ($1:4$) delivers dramatic speedups at the cost of noticeable quality degradation. This suggests that a trade-off exists between the degree of sparsification and output quality, with an optimal balance likely achieved through a compromise between 2:4 and 1:4 configurations. 
The findings are particularly relevant for production environments. Our data-driven sparsification strategy is specifically engineered for commercial models' parameter distributions, with targeted optimization for high-frequency, business-critical features as elaborated in \secref{sec:solution}. 
While generalized applications warrant further investigation, this tailored methodology demonstrates superior performance in industrial deployments where practical constraints are paramount.


\begin{table}[!htbp]
\caption{Ablation studies of quantization and sparsification}
\label{ablation}
\vspace{-2ex}
\begin{tabular}{lcccc} 
\toprule
Technique & Latency & Meteor & BLEU & Speedup \\
\midrule
Baseline (FP16) & 6.6 & 0.6345 & 0.4247 & $\times$1.00 \\
Quantization & 4.8 & 0.6283 & 0.4178 & $\times$1.37 \\
Sparsification(2:4) & 5.3 & 0.6260 & 0.4161 & $\times$1.25 \\
Sparsification(1:4) & 4.6 & 0.5549 & 0.3476 & $\times$1.43 \\
Sparse(2:4) + Quant & 4.0 & 0.6195 & 0.4103 & $\times$1.65 \\
Sparse(1:4) + Quant & 3.5 & 0.5446 & 0.3369 & $\times$1.89 \\
Sparse(Mix) + Quant & 3.7 & 0.6127 & 0.4038 & $\times$1.78 \\
\bottomrule
\end{tabular}
\end{table}
\vspace{-2ex}

\section{Conclusion}\label{sec:conclusion}

This study presents the \TechName framework, an innovative solution to the generative targeting problem in search advertising, optimizing inference efficiency without sacrificing precision. We address this challenge with two key strategies: model compression and prefix tree parallel verification. Our extensive experiments show that \TechName outperforms previous LLM-based advertising systems in real-time delivery settings, demonstrating its effectiveness and potential for practical deployment in dynamic advertising environments.
However, the framework's current design is primarily suited for commercial advertising targeting scenarios, as its optimization strategies (quantization, sparsification, and business-oriented improvements) are tailored to the unique requirements of this domain. Applications in other contexts may require further adaptation to address differences in data patterns, objectives, and constraints. Future research will focus on integrating adaptive algorithms and reinforcement learning to further enhance efficiency, scalability, and adaptability, aiming to refine the balance between optimization and user experience in modern advertising.

\bibliographystyle{ACM-Reference-Format}
\bibliography{8_ref}

\end{document}